\documentclass[a4paper,fleqn]{cas-dc}
\usepackage[numbers,sort&compress]{natbib}
\usepackage{microtype}
\usepackage{tabularx}
\usepackage{placeins}
\usepackage{xurl}
\usepackage{lastpage}
\AtBeginDocument{}

\ExplSyntaxOn
\RenewDocumentCommand \firstname {} { \textcolor{black}{\seq_use:Nn \l_stm_au_seq { ~ }} }
\ExplSyntaxOff
\newcolumntype{Y}{>{\centering\arraybackslash}X}
\newcolumntype{Z}{>{\raggedright\arraybackslash}X}
\RenewDocumentEnvironment{Abstract}{o}
{%
  \IfNoValueF{#1}{\renewcommand{\abstractname}{#1}}%
  \setlength{\parindent}{0pt}%
  \noindent\abstractname\par
  \noindent\rule{\textwidth}{.2pt}\par
  \footnotesize\noindent\ignorespaces
}
{\par}

\ExplSyntaxOn
\cs_set:Npn \__first_footerline: { \mbox{} }
\ExplSyntaxOff

\begin{document}
\let\WriteBookmarks\relax
\def\floatpagepagefraction{1}
\def\textpagefraction{.001}
\shorttitle{Resilient SDLs through telerobotic interventions}
\shortauthors{Z. Wu et~al.}
\title[mode=title]{Toward Resilient Self-Driving Laboratories through Telerobotic Interventions}
\author[1]{Ziheng Wu}
\cormark[1]
\ead{wu57@llnl.gov}
\author[2,4]{Adrian Vecina Tercero}
\author[1]{Andre A. Fatehi}
\author[2]{Abhik Sarkar}
\author[3]{Brian Giera}
\author[3]{Eric B. Duoss}
\author[1]{Christopher M. Spadaccini}
\author[4]{Nikhil Deshpande}
\author[1]{Aldair E. Gongora}
\cormark[1]
\ead{gongora1@llnl.gov}
\affiliation[1]{organization={Materials Engineering Division, Lawrence Livermore National Laboratory},
  city={Livermore}, state={CA}, postcode={94550}}
\affiliation[2]{organization={Center for Applied Scientific Computing, Lawrence Livermore National Laboratory},
  city={Livermore}, state={CA}, postcode={94550}}
\affiliation[3]{organization={Engineering Directorate, Lawrence Livermore National Laboratory},
  city={Livermore}, state={CA}, postcode={94550}}
\affiliation[4]{organization={School of Computer Science, University of Nottingham},
  city={Nottingham}, country={UK}}
\cortext[1]{Corresponding authors.}

\begin{abstract}
Self-driving laboratories (SDLs) are accelerating scientific research by integrating robotics, automated experimentation, and machine learning to design and execute experiments autonomously. Fields such as biology and chemistry have shown early success in adopting these systems. However, the increasing complexity of SDLs has heightened their susceptibility to unanticipated faults within automated workflows. These disruptions underscore the need for resilient mechanisms that can effectively address faults without significantly compromising operational efficiency. In this study, we investigate the integration of telerobotic interventions as a solution to enhance the resilience of SDLs against unexpected failures. To address the challenges of fault recovery in SDLs, we developed a teleoperation interface that enables human operators to intervene and resolve laboratory disruptions during automated task executions. The teleoperation interface was designed to allow precise control of a six-degree-of-freedom (DOF) robotic arm (UR3e) using a 3D mouse (3Dconnexion SpaceMouse Pro). Two velocity-based controller strategies were evaluated: (i) a coupled mode, where all six DOFs are mapped to the central knob of the 3D mouse, and (ii) a decoupled mode, which separates translational and rotational motions into distinct controls. To evaluate the framework, teleoperators performed interventions across three distinct SDL fault scenarios. In the first study, comparative trials revealed that the decoupled controller was associated with shorter completion time and reduced rotational motion during teleoperation. In a second study, teleoperators performed an aiming task with and without guidance-providing reference paths. Users were seen to be more target-driven and risk-tolerant without guidance but did show deviations in behavior when presented with different guidance strategies. These findings suggest that controller configuration and reference-path design can influence operator performance during SDL fault recovery tasks, providing insights into optimizing teleoperation interfaces. This work demonstrates the potential feasibility of integrating teleoperation as a human-in-the-loop recovery mechanism for SDL workflows, combining the precision of robotics with the adaptability and problem-solving capabilities of human operators. Furthermore, these findings provide a foundation for future studies in shared autonomy and semi-autonomous intervention.
\end{abstract}
\AtEndEnvironment{Abstract}{%
  \par
}
\maketitle

\section{Introduction}\label{introduction}

Self-driving laboratories (SDLs) have emerged as a transformative approach to accelerating scientific research in chemistry, materials science, and biology by integrating robotics, laboratory automation, and machine learning~\cite{ref01}. Due to modern developments in robotics and artificial intelligence, researchers envision SDLs as engines to accelerate discovery where research platforms are capable of autonomously designing and executing experiments with knowledge from subject matter experts and domain scientists. Interestingly, the conceptual foundation of SDLs trace back to the 1960s. Notably, the DENDRAL project~\cite{ref02} developed the first software with predefined chemistry rules to predict chemical structures using mass spectrometry data. By the 1980s, laboratory robotic systems were gaining traction in analytical chemistry, giving rise to early prototypes of what are now regarded as SDLs~\cite{ref03,ref04,ref05}.

With the continued rise of SDL technologies across diverse scientific domains in recent decades, a formal autonomy classification framework has been established to characterize SDL technology. The framework parallels the automation levels for self-driving vehicles by the Society of Automotive Engineers (SAE)~\cite{ref06}. Specifically, SDLs have been categorized into five levels of autonomy~\cite{ref07,ref08,ref09,ref10}: (I) assisted operation, (II) partial autonomy, (III) conditional autonomy, (IV) high autonomy, and (V) full autonomy. Each successive level reflects an increase in system complexity and integration across both software intelligence and hardware control. A few platforms, such as Adam~\cite{ref11}, the BEAR~\cite{ref12}, ADA~\cite{ref13} A-Lab~\cite{ref14}, and MicroCycle~\cite{ref15}, have achieved level IV autonomy, wherein experimental design, execution, and analysis proceed with minimal human intervention.

Progressing toward higher levels of autonomy inherently exposes SDLs to greater demands on reliability, robustness, and adaptability, particularly as the architectural system complexity of SDLs continues to increase. It will require systems that not only operate without human supervision but also recognize, diagnose, and correct their own errors in real time. However, current SDL architectures often presuppose near-perfect automation, relying on tightly coupled hardware-software infrastructures. For example, six-axis robotic arms are frequently hard-coded within highly constrained setups that depend on precise fixturing and predefined placement of laboratory objects~\cite{ref07,ref09}. In practice, real-world experimental environments remain inherently variable and prone to disruptions such as robotic misalignments, hardware of software malfunctions, and variable human-system interactions. Beyond degrading performance, such failures can introduce unsafe conditions, including unintended robot motion, material spills, or damage to laboratory infrastructure, thereby posing risks to both human operators and experimental assets. These perturbations frequently interrupt workflows and can lead to cascading system failures, exposing the limited \emph{system} \emph{resilience} of current SDL architectures~\cite{ref16,ref17}. \emph{System resilience}, as defined by National Institute of Standards and Technology (NIST)~\cite{ref18}, refers to the ability to continue operating under adverse conditions or stress and recover promptly to support mission needs. Advancing SDL autonomy will therefore require a transition from rigid, deterministic architectures toward adaptive frameworks capable of autonomous fault detection, dynamic recovery, and human-in-the-loop intervention to reduce workflow interruptions and maintain safe operation. Embedding resilience as a foundational design principle will be essential for developing the next generation of SDLs that are capable of sustained autonomous operation.

To address the challenge of enhancing system resilience in SDLs, we consider robotic teleoperation as a promising first step. Teleoperation enables human operators to directly intervene in real time, extending human cognition and dexterity into robotic systems while preserving the precision, repeatability, and stability of automation~\cite{ref19}. Importantly, teleoperation provides a mechanism for mitigating safety risks by allowing humans to respond immediately to anomalous or hazardous system states without direct physical exposure. This is particularly valuable in uncertain or dynamic environments, where autonomous systems may struggle to respond effectively~\cite{ref20}. Fortunately, the concept is well established in fields, such as space exploration, telemedicine, and nuclear operations, that demand reliability under constrained or hazardous conditions~\cite{ref19,ref21}. In these domains, teleoperation has served as a bridge between human judgement and machine precision. In laboratory contexts, researchers have also demonstrated the use of teleoperation in a glovebox~\cite{ref22,ref23} to mitigate the risk of human exposure to hazardous materials. Building on these efforts, teleoperation represents an underexplored yet powerful avenue for developing SDLs that can recover from unexpected disruptions. Furthermore, recent advances in teleoperation, ranging from optimal control strategies and haptic feedback to immersive virtual and mixed-reality interfaces, have significantly enhanced operator performance and situational awareness through user-centered design~\cite{ref24,ref25,ref26}. Collectively, these developments position teleoperation as one of the practical pathways toward more resilient SDLs.

Complementing these advances, on-going research has focused on improving the intuitiveness and safety of teleoperation systems through assistive control and guidance strategies~\cite{ref27}. Many platforms utilize virtual fixtures to guide users along the task-specific path and prevents any unintended collision~\cite{ref28}. Compliance-based control schemes, such as those proposed by Bischof \emph{et al.}~\cite{ref29}, combine surface-following and constraint enforcement to implement forbidden regions, dynamic boundaries, and reference path guidance. Nagano \emph{et al.}~\cite{ref30} designed a guidance system that provides haptic feedback based on trajectory similarity to improve maneuvering performance. Their approach predicts a reference trajectory based on the posture trajectory of the local control interface approaches with predictive displays and semi-autonomous control frameworks further improve user performance by reducing cognitive workload and aid with collision avoidance~\cite{ref31,ref32}. Yet, few studies evaluated the selection of these reference paths. Meanwhile, Gottardi \emph{et al.}~\cite{ref33} proposed a shared-control framework that assists operators to navigate between static and dynamic obstacles. Using the improved potential field method, their system dynamically generates escape waypoints which serve as attractors to the teleoperation path. Collectively, these contributions highlight a robust foundation for integration of advanced teleoperation methodologies into SDLs, paving the way for hybrid systems that blend autonomy with adaptive human oversight.

\subsection{Objectives \& Contributions}\label{objectives-contributions}

Current SDLs often lack teleoperation-assisted recovery mechanisms that allow humans to safely and efficiently intervene when unexpected faults arise. The goal of this work is not to introduce a fundamentally new teleoperation algorithm, but rather to investigate how teleoperation can be leveraged to strengthen the system resilience of SDLs by enabling human-guided fault recovery.

We first discuss the development of telerobotic capabilities, building upon prior advances in teleoperation interface design and control, to create an intuitive and adaptive human--machine collaboration environment. The system centers on a six-axis robotic arm controlled via a 3D input device, employing velocity-based motion mapping for precise and responsive real-time manipulation. Among available teleoperation interfaces, the 3D SpaceMouse offers a cost-effective and widely adopted solution for achieving fine-grained spatial control~\cite{ref34,ref35,ref36}. We then evaluate control strategies for telerobotic operation through a preliminary user study, quantifying the trade-offs among different velocity-based control modes and their influence on user performance during representative SDL fault recovery tasks.

To further assess how teleoperation can improve operator performance, we implement another user study to investigate the effect of reference-path guidance, analyzing how trajectory design affects teleoperator performance. From selected teleoperated trajectories, we retrospectively analyze collision and singularity risk using three quantitative indices that characterize controllability, system responsiveness, and motion stability, providing a principled basis for mitigating self-entanglement and improving task efficiency.

Finally, we integrate these telerobotic capabilities into a real-world SDL setup for a qualitative demonstration, illustrating how human-in-the-loop fault recovery could significantly improve laboratory uptime and operational robustness. Collectively, these studies establish a quantitative, qualitative, and experimental foundation for teleoperation-assisted fault recovery, marking a step toward resilient, adaptive, and fault-tolerant SDLs capable of sustained operation under real-world uncertainty.

\section{Telerobotic System Overview for Applications in Self-driving Labs (SDLs)}\label{telerobotic-system-overview-for-applications-in-self-driving-labs-sdls}

To enhance the resilience of self-driving laboratories (SDLs) and enable real-time human intervention during faults, we developed a teleoperation framework that augments autonomous laboratory workflows with operator-in-the-loop control (Figure~\ref{fig:1}A). Drawing on prior advances in controller design and teleoperation ergonomics, the framework provides an intuitive and adaptive interface that allows human operators---including non-experts---to safely and efficiently assist with recovery tasks.

\subsection{Teleoperation Setup}\label{teleoperation-setup}

The design of the teleoperation system was guided by prior research emphasizing the importance of intuitive controller--interface configurations for fault recovery and precision manipulation. Han and Yang~\cite{ref37} demonstrated that appropriate selection between position and rate control modes can significantly enhance teleoperator performance, while other studies have shown that interface design directly influences operational efficiency and safety~\cite{ref38}. Dhat \emph{et al}.~\cite{ref39} optimized SpaceMouse-driven teleoperation by addressing issues of high sensitivity, dimensional coupling, and limited feedback through dimensional weighting, axis-aligned motion constraints, and input magnification.

\begin{figure*}[pos=tp]
\centering
\includegraphics[width=1.0\textwidth,height=0.70\textheight,keepaspectratio]{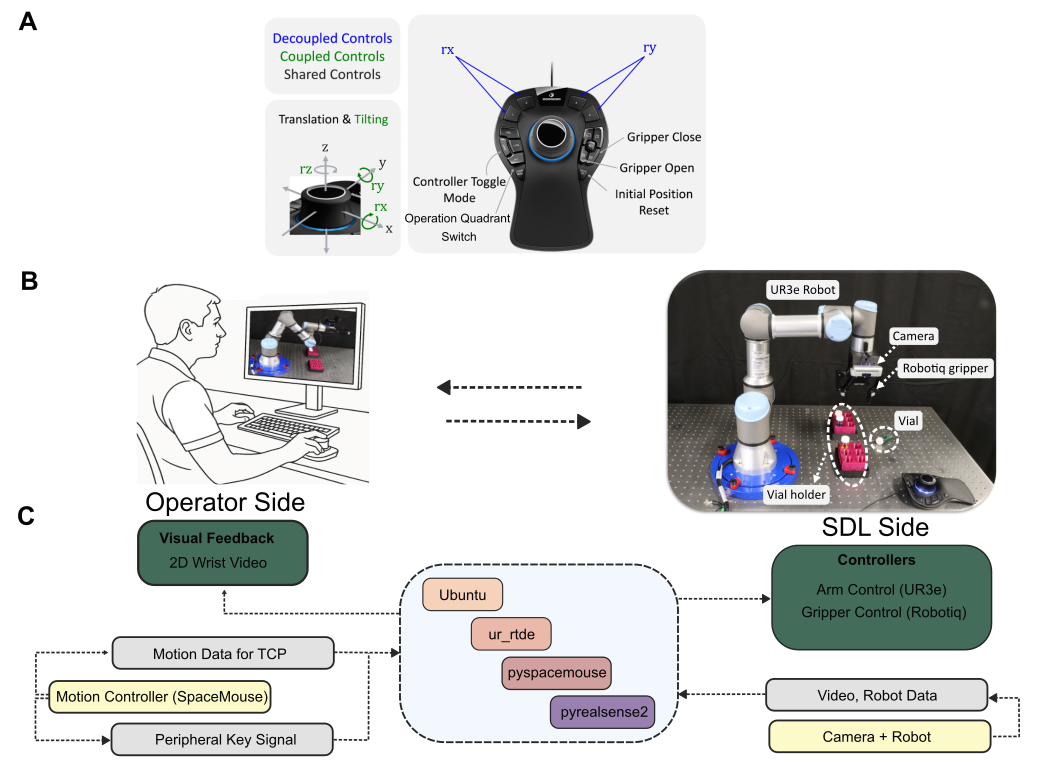}
\caption{Schema shows (A) the configuration of the SpaceMouse, and the system overviews of (B) the hardware and (C) the software of the SDL robotic teleoperation setup. Keys on SpaceMouse are annotated in green for the coupled control mode and annotated in blue for the decoupled mode.}
\label{fig:1}
\end{figure*}

Building on these insights, the system architecture integrates the traditional teleoperation setup, having an \emph{operator side}, the \emph{SDL side}, and the communication bridge in between. The following components:

\begin{enumerate}
\def\labelenumi{(\arabic{enumi})}
\item
  A six-axis robotic arm (Universal Robots UR3e) and a vision module (Intel RealSense D435 or Logitech C920e) represent the SDL-side setup. The vision module is mounted at the end-effector wrist location of robot, as seen in Figure~\ref{fig:1}A (right panel). The SDL side includes vials \& vial holders to undertake representative tasks in lab automation, \emph{e.g.}, pick-n-place.
\item
  For the operator side, a 3D input device (3Dconnexion SpaceMouse\textregistered{} Pro), which serves as the primary input device, offering six degrees of freedom (DOFs) and fifteen programmable keys. Through these inputs, users can toggle between control modes, operate the gripper, and reset the SDL robot's position. During teleoperation, the robot wrist-mounted camera provides a live visual feed to the operator through a workstation monitor, giving situational awareness of the immediate robot workspace and allowing operators to perform targeting and alignment tasks in real time.
\item
  A Linux workstation coordinates the communication \& command exchange between the user and the SDL setup. The SpaceMouse connects to the workstation via USB, while the UR3e communicates with it through the RTDE protocol over TCP/IP using the open source \emph{ur\_rtde} Python package (Figure~\ref{fig:1}B).
\end{enumerate}

During teleoperation, the SDL operates in remote mode, executing external scripts that map SpaceMouse inputs to the robot tool-center point (TCP) position and orientation. Together, these components form an adaptive telerobotic platform capable of extending SDL autonomy by embedding human perception and decision-making within the control loop.

\subsection{Telerobotic Control Mapping}\label{telerobotic-control-mapping}

Our implementation employs velocity-based motion mapping between the SpaceMouse and the robotic arm to achieve precise, stable, and responsive manipulation. This configuration enables the operator to intuitively guide the robot through complex recovery tasks, providing a foundation for subsequent studies on testing control strategies, the reference-path guidance approach, collision-risk mitigation, as well as real-world deployment within SDL environments.

For the control strategies of telerobotic operation, we first consider the mapping of all six DOFs of the TCP to the SpaceMouse controller. We refer to this strategy as \emph{coupled} mode, since robotic motions are fully manipulated by the center knob on the SpaceMouse controller. We also consider a second control strategy, namely \emph{decoupled} mode, which maps only four DOFs \((x,\ y,z,\ rz)\) to the center knob while the pitch and yaw \(( \pm rx,\ \  \pm ry)\) are mapped to four separate peripheral keys. The velocity-based controller strategy was selected over a position-based controller strategy due to the non-homothetic nature of the coupling of the robotic arm and the SpaceMouse~\cite{ref40}. Our approach utilizes unilateral teleoperation architecture. The motion commands from the SpaceMouse are sampled synchronously and sent to the robot at a fixed rate of 500 Hz to ensure responsiveness and stability. The inputs from the SpaceMouse are mapped directly to the Cartesian velocity of the end effector:

\begin{equation}\label{eq:velocity-mapping}
\vec{V}_{tcp} =
\begin{bmatrix}v_x\\v_y\\v_z\\\omega_x\\\omega_y\\\omega_z\end{bmatrix}
= \begin{bmatrix}s_v u_x\\s_v u_y\\s_v u_z\\s_\omega r_x\\s_\omega r_y\\s_\omega r_z\end{bmatrix}.
\end{equation}

where \(s_{v}\) and \(s_{\omega}\) are the scaling factors for the translational and the angular velocity, \({\overrightarrow{u}}_{t} = \left\lbrack u_{x},\ u_{y},u_{z} \right\rbrack^{T}\) is the translational input vector, and \({\overrightarrow{u}}_{r} = \left\lbrack r_{x},r_{y},r_{z} \right\rbrack^{T}\) is the rotational input vector. The system allows the users to adjust the acceleration and scaling factors, which can be key to smooth manipulation. We used an acceleration of 1.2 m/s\textsuperscript{2}, \(s_v = 0.1\), and \(s_\omega = 0.5\) in our setup, which were empirically determined and deemed suitable for our study.

The rest of the peripheral keys on the SpaceMouse are identical in both controller modes, as shown in Figure~\ref{fig:1}, performing functions of opening and closing the gripper, switching control modes, resetting the arm position, and switching the operation quadrants for solving the coordinate-frame inversion problem. These commands are communicated within the same thread as the motion commands.

A separate thread handles the data logging. It monitors the keypress events that trigger the start and the stop of the recording process. Once triggered, the system starts video recording at the wrist-mounted camera and environment camera at 30 fps; meanwhile, it documents all the joint positions, the tool center point positions, the gripper state of the robotic arm at a frequency of 20 Hz.

\subsection{Statistical Analysis}\label{statistical-analysis}

Statistical analyses were performed for the studies conducted using the teleoperation setup. We utilized participant-level aggregated metrics to account for repeated trials from the same operator.

For the control strategy study, paired comparisons between coupled and decoupled modes were conducted using the Wilcoxon signed-rank test, with paired t-tests additionally reported as a secondary reference. Effect sizes were quantified using Cohen's dz based on paired participant differences. Bootstrap 95\% confidence intervals (CI) were computed using 10,000 resamples. Success rates were analyzed using Fisher's exact test with Wilson confidence intervals for binomial proportions. Because of the limited participant count and exploratory nature of the study, statistical analyses are intended to characterize performance trends and uncertainty rather than establish definitive generalizable conclusions. Additional statistical summaries are provided in supplementary materials Table S1.

For the reference-path guidance study, comparisons between unguided and guided conditions were interpreted as exploratory because guided trials were conducted after unguided trials, introducing potential practice effects. Additional statistical summaries for the guidance study are provided in Tables S2 and S3.

\section{Evaluating Control Strategies for Telerobotic Operation}\label{evaluating-control-strategies-for-telerobotic-operation}

\subsection{Study Design}\label{study-design}

\begin{figure*}[pos=tp]
\centering
\includegraphics[width=1.0\textwidth,height=0.70\textheight,keepaspectratio]{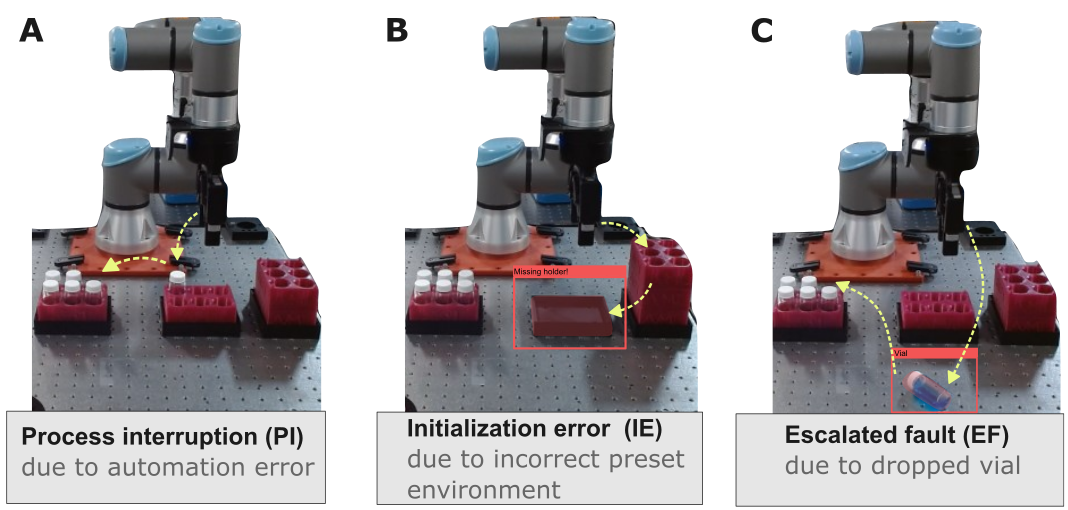}
\caption{Example robotic trajectories for resolving three simulated SDL fault scenarios: (A) process interruption (PI) which represents a mid-process failure halting the automated workflow, (B) initialization error (IE) represents a pre-start failure caused by improper system or workflow setup, (C) escalated fault (EF) represents a severe operational error, such as end-effector malfunction resulting in a dropped object.}
\label{fig:2}
\end{figure*}

To evaluate and compare the performance of the two strategies, \emph{i.e.}, the \emph{coupled} mode vs. the \emph{decoupled} mode, we designed a study comprising three simulated SDL fault scenarios, namely process interruption (PI), initialization error (IE), and escalated fault (EF) (Figure~\ref{fig:2}). The simulated faults were designed to represent a range of common errors in laboratory automation and autonomous experimentation. In each scenario, the teleoperators were tasked with using the robotic arm to resolve the fault and restore the system to its nominal state of operation. Each controller mode was tested independently in separate trials.

\begin{enumerate}
\def\labelenumi{\arabic{enumi}.}
\item
  \emph{Process interruption (PI)} represents a mid-process failure in which an automated task halts unexpectedly at an intermediate step, requiring human intervention to complete the action and return the robot to a recoverable configuration (Figure~\ref{fig:2}A). Here, the teleoperators were tasked with guiding the robot to continue transferring a glass vial from one vial holder to another vial holder, restoring continuity to the interrupted process.

\item
  \emph{Initialization error (IE)} represents a pre-start failure caused by improper system or workflow setup (Figure~\ref{fig:2}B). In this scenario, a vial holder was intentionally removed, creating an incomplete configuration that would otherwise trigger an error during automated execution. Here, the teleoperators were required to identify the missing component, retrieve it, and correctly position it at the designated location to re-enable the operation.

\item
  \emph{Escalated fault (EF)} represents a severe operational error, such as end-effector malfunction resulting in a dropped object (Figure~\ref{fig:2}C). In this scenario, a glass vial was released during transfer, representing a critical failure that would halt the experimental workflow. Teleoperators were required to reorient the arm at a tilted angle, grasp the fallen vial, and deliver it to the designated holder. In this task, the teleoperators needed to precisely manipulate the robotic arm across all six DOFs.
\end{enumerate}

Collectively, the three scenarios reflect representative categories of faults encountered in laboratory automation workflows or SDLs. Process interruptions (PI) often arise from communication or orchestration failures among interconnected instruments~\cite{ref41}. Initialization errors (IE) mirror setup inconsistencies that emerge from the high configuration and setup demands of automated workflows. Escalated faults (EF) capture the rare but high-impact cases where physical manipulation errors, such as dropped samples or objects, necessitate direct and immediate human intervention. Beyond their functional relevance, these three scenarios were also designed to impose increasing kinematic and cognitive complexity, where PI \textless{} IE \textless{} EF. Completing PI requires only two planar translations where resolving IE adds roll and all translational motions. Furthermore, EF requires engaging all six DOFs for adequate manipulations which increases the risk of singularities and collisions. Together, these cases provide a systematic approach for evaluating teleoperator performance across escalating levels of difficulty and constraints, while emulating real-world fault conditions in SDL environments. While these scenarios may not encompass the full spectrum of faults that may arise in complex autonomous laboratories, this work represents an important step toward bridging fully autonomous operations with practical human-in-the-loop interventions, leveraging both the precision of automation and the adaptability of human teleoperation.

The performance of the two control strategies was evaluated using data from six participant teleoperators who collectively completed 86 teleoperation trials which included 28 runs each for PI and EF scenarios, and 30 runs for IE scenarios. The six male participants have limited prior teleoperation experience and no exposure to our teleoperation setup before. The coupled and decoupled controller modes were equally represented across all assessments.

\subsection{Study Results \& Discussion}\label{study-results-discussion}

\begin{figure*}[pos=tp]
\centering
\includegraphics[width=1.0\textwidth,height=0.70\textheight,keepaspectratio]{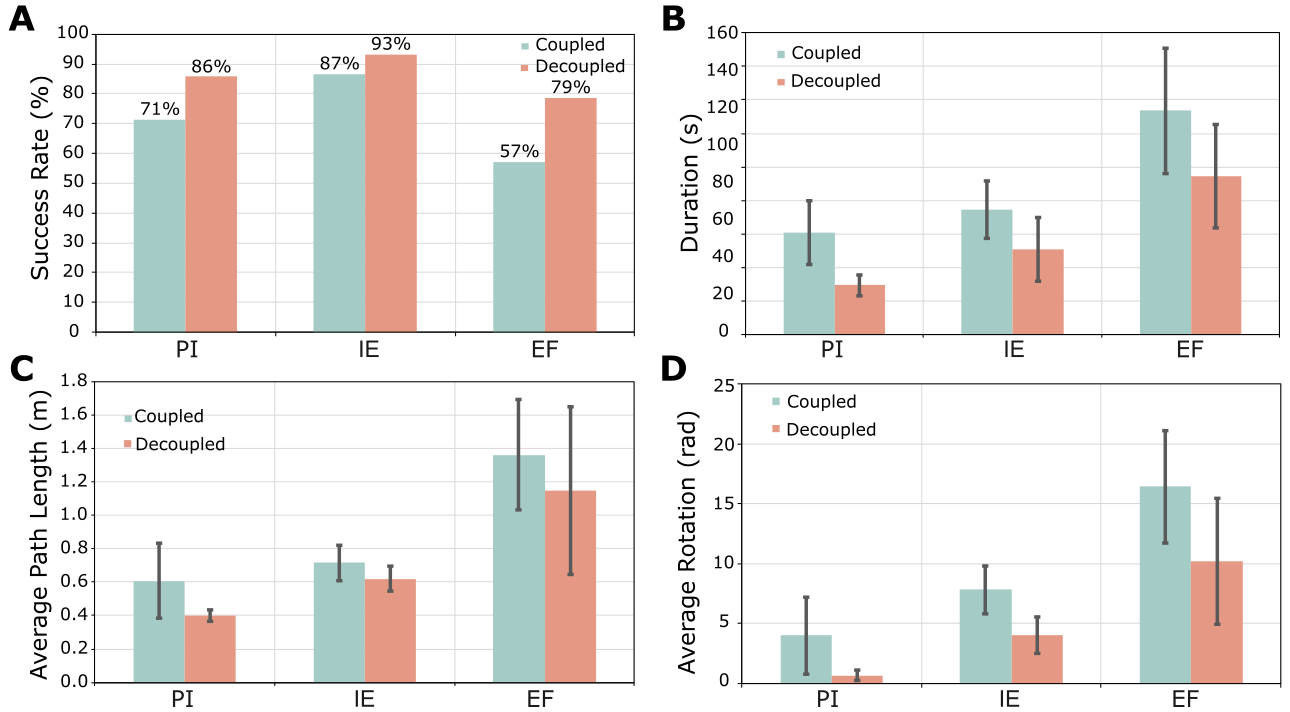}
\caption{Assessing capabilities of the coupled and the decoupled telerobotic controller modes in handling three simulated SDL fault scenarios: process interruption (PI), initialization error (IE), escalated fault (EF). The operations are quantified in (A) success rate, (B) average rotation, (C) duration, and (D) average path length. The error bars represent one standard deviation. Note that only successful cases are included in the calculations of the duration, the average path length, and the average rotation.}
\label{fig:3}
\end{figure*}

Overall, the decoupled control mode consistently outperformed the coupled mode across all three fault scenarios, achieving a higher success rate, faster task completion times, and more efficient motion trajectories (Figure~\ref{fig:3}A--D).

The success rate (Figure~\ref{fig:3}A), defined as the proportion of trials in which operators successfully resolve the fault. A trial was classified as successful if the operator completed the designated recovery task without collision, force-triggered interruption, overextension of robotic arm or workspace violation. The success rate revealed the most pronounced differences in tasks requiring higher precision. For example, in the PI scenario, success improved from 71\% to 86\% under the decoupled mode. This task requires lifting the vial vertically from a snug holder while maintaining strict alignment between the vial and the holder axes where even minor wrist rotations can induce excessive friction and trigger the robot's force protection mechanism. By separating translational and rotational inputs, the decoupled mode isolates pitch and yaw control from the SpaceMouse's central knob, reducing motion interference and improving path stability. Similar trends were observed in the IE and EF scenarios, where success rates increased by 6\% and 22\%, respectively.

Reducing coupling between translation and rotation not only improved path stability but also minimized redundant motion (Figure~\ref{fig:3}). The decoupled mode was associated with reduced rotational motion relative to the coupled mode, with a mean paired difference of -3.99 rad (bootstrap 95\% CI: -6.59 to -1.22 rad). Although the Wilcoxon comparison did not reach conventional statistical significance (p = 0.094), the observed effect size remained large (Cohen's dz = -1.08), suggesting a meaningful reduction in unnecessary rotational adjustments during teleoperation. Across all three fault scenarios, the PI task showed the most dramatic reduction in angular displacement. This reduction in average rotation is attributed to the strategy selection in the PI scenario which reduces redundant rotations. Efficiency metrics further reinforce this improvement where the decoupled mode consistently produced shorter average path lengths and reduced task durations (Figure~\ref{fig:3}B and Figure~\ref{fig:3}C). Across paired participant means (n = 6), the decoupled mode was associated with reduced task completion time relative to the coupled mode, with a mean duration difference of -25.6 s (decoupled \ensuremath{-} coupled; bootstrap 95\% CI: -36.6 to -14.2 s). Wilcoxon signed-rank testing suggested shorter completion times under the decoupled condition (p = 0.031), with a large paired effect size (Cohen's dz = -1.60). These findings suggest that the decoupled strategy improves operation efficiency through a more cognitively ergonomic approach for human operators, mitigating cross-axis interference and reducing the need for trajectory correction, thereby streamlining teleoperation during fault recovery in SDLs.

\section{Investigating the Effect of Reference Path Guidance on User Performance}\label{investigating-the-effect-of-reference-path-guidance-on-user-performance}

Building on the findings from the previous section, which showed that controller configuration can affect task efficiency and operator performance, we next investigate how guidance in the form of reference paths may affect the performance of teleoperators. When anomalies in an SDL occur, teleoperators must diagnose the root cause, identify obstacles, and construct a collision-free recovery trajectory, often under time pressure and mission-critical constraints. Under these conditions, we hypothesize operators may benefit from guidance using pre-computed near-optimal reference path~\cite{ref28}. For example, a favorable reference path for robotic arm navigation may consider the minimization of the path length. While the derived path may be simple for automatic robotic execution using inverse kinematics, the derived path may impose additional cognitive demands on human operators who must disentangle multi-DOF motions and reproduce them through a robotic controller. Additionally, the derived path may also be a suboptimal solution for the task due to collision or singularity risk. To investigate the effect of reference-path guidance on teleoperation performance, we evaluated user execution of representative tasks under two conditions, namely with and without reference-path guidance. For the sake of simplicity and clarity, the reference paths were conveyed to the teleoperator through direct demonstration in our studies.

\subsection{Study Design}\label{study-design-1}

\begin{figure*}[pos=tp]
\centering
\includegraphics[width=1.0\textwidth,height=0.70\textheight,keepaspectratio]{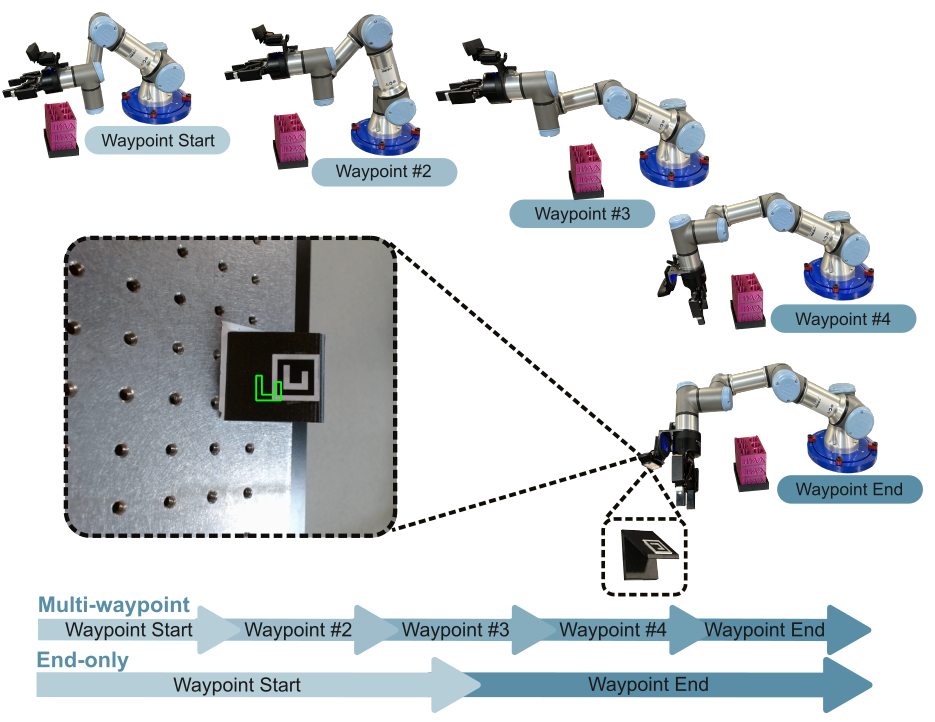}
\caption{A schematic shows the arm configurations of the multi-waypoint and the end-only reference paths at each waypoint. The magnified image highlights the goal of the operation which is to align the virtual and the physical markers.}
\label{fig:4}
\end{figure*}

In practical SDL environments, operators performing intervention may need to navigate around constrained laboratory equipment, sample containers, tubing, or hazardous regions. Under these conditions, guidance strategies may influence recovery efficiency and collision avoidance behavior. Here, participants were asked to teleoperate the UR3e robotic arm to navigate around obstacles while moving between two spatial targets. The objective of the task was to align two markers: a virtual marker displayed in the video stream from the robot wrist-mounted camera (Figure~\ref{fig:1}A) and a physical ArUco marker~\cite{ref42} placed on the platform (Figure~\ref{fig:4}). Figure~\ref{fig:4} illustrates the robot's configuration at each waypoint along the two reference paths. Based on the superior controllability demonstrated in Section~\ref{evaluating-control-strategies-for-telerobotic-operation}, all experiments were conducted using the decoupled controller mode. Participants completed the task under three experimental conditions designed to examine different forms of guidance and their impact on teleoperation performance.

\begin{enumerate}
\def\labelenumi{\arabic{enumi}.}
\item
  \emph{Unguided condition}: This baseline condition captures the natural strategies and learning trajectories of novice operators without reference-path assistance. These assessments are expected to reveal the underlying natural tendencies in human telemanipulation approaches.
\item
  \emph{End-only} \emph{reference path}: This condition provides a direct trajectory from the start to the end waypoint without intermediate step. This assessment introduces higher kinematic complexity due to the simultaneous, nonlinear joint motions required in both position and orientation space.
\item
  \emph{Multi-waypoint} \emph{reference path}: This condition introduces three intermediate waypoints that decompose the overall motion into four sequentially simpler segments, consisting primarily of linear movements in the z and x directions followed by pitch and yaw adjustments.
\end{enumerate}

Participants were briefed on the details of the two reference paths prior to the trials. The \emph{end-only reference path,} derived via inversive kinematics, was designed to prioritize efficiency and represents the shortest feasible trajectory for this task (375.04 mm). In contrast, the multi-waypoint reference path (755.10 mm) was intentionally constructed to (i) help inexperienced operators better understand the correspondence between controller inputs and robot motion, and (ii) reduce collision risk by initiating the motion with an upward translation in \emph{z} to raise the wrist, thereby avoiding potential contact with the vial holder.

Eight teleoperators (six males and two females) participated and each was asked to perform 12 trials resulting in total of 96 datasets. The moderator concealed the two reference paths from the operators for the first six runs, which are labeled as \emph{unguided} condition. Before trial \#7, each reference path was shown three times to the operators, who were then asked to imitate the end-only reference path in trial \#7, \#9, and \#11, and the multi-waypoint reference path in trial \#8, \#10, and \#12 to achieve the same goal.

\subsection{Study Results \& Discussion}\label{study-results-discussion-1}

\begin{figure*}[pos=tp]
\centering
\includegraphics[width=1.0\textwidth,height=0.70\textheight,keepaspectratio]{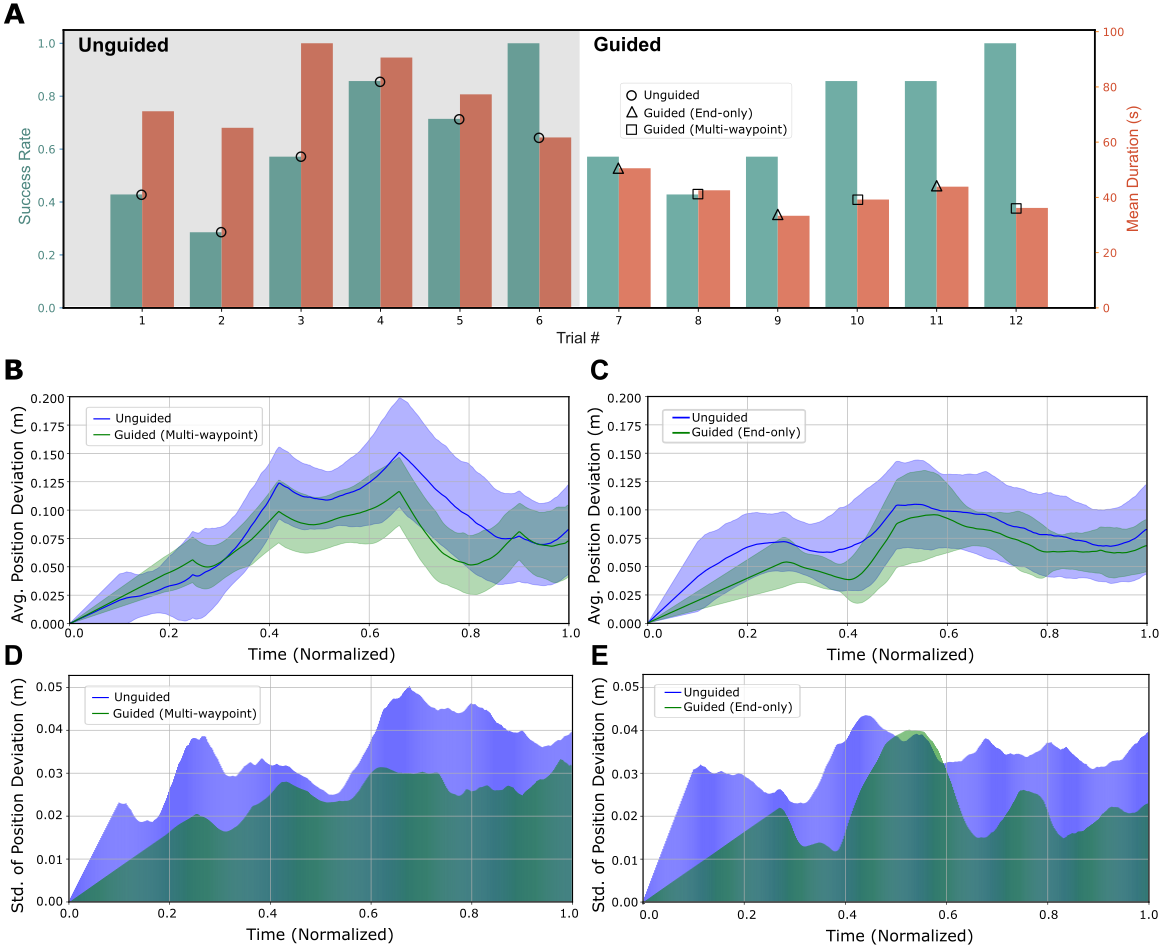}
\caption{Evaluating how practice and guidance affect users' performance and measuring the differences and similarities between the teleoperation paths and the reference paths. Quantifying (A) the average success rate and the mean duration as a function of trial number,  (B) the position deviations and (D) the corresponding variations by comparing the unguided and multi-waypoint guided paths with the multi-waypoint reference, as well as (C) the position deviations and (E) the corresponding variations by comparing the unguided and end-only guided paths with the end-only reference. The solid curves in (B) and (C) track the average deviations, and the shaded areas present \ensuremath{\pm} 1 standard deviation which is represented in (D) and (E). Note that only successful trials contribute to the deviation and the duration calculation.}
\label{fig:5}
\end{figure*}

The success rate (Figure~\ref{fig:5}A) is defined as the proportion of trials in which operators successfully align the virtual and the physical markers. The success and failure of the alignment task was judged manually by the same expert who directly monitored the camera feed. With minimal prior exposure to teleoperation, the participants achieved a \textless{} 50\% success rate in the initial two runs. Most failures were caused by overextension and singularity. Users reported that their primary focus was on collision avoidance and operation speed. From trial \#3 to \#6, users started to recognize the risk of overextension and singularity. They became more cautious, as reflected by the increased mean duration and the improved success rate. At trial \#6, all users solved the challenge without any failures, resulting in a 100\% success rate.

When a reference path is provided, we observed a similar trend in improved success rate and performance. However, failure shortly reemerged at trial \#7. This early dip is expected, as users must adopt new strategies for task execution and potentially over-prioritize path imitation rather than risk avoidance. Yet, only after two additional trials, users achieved an 85\% and 100\% success rate when imitating the end-only path and the multi-waypoint path, respectively. Exploratory paired comparisons between the multi-waypoint guided condition and the unguided condition suggested reduced task duration under guided operation, with a mean paired difference of -40.1 s (guided multi-waypoint \ensuremath{-} unguided; paired n = 7; bootstrap 95\% CI: -55.9 to -26.3 s; Wilcoxon p = 0.016). A similar trend was observed for path length, where multi-waypoint guidance was associated with shorter teleoperation trajectories relative to unguided operation (mean paired difference = -218 mm; bootstrap 95\% CI: -365 to -81 mm; Wilcoxon p = 0.047). While we observed improved final success rates, faster operation and shorter paths, these alone do not necessarily mean that the end-only or the multi-waypoint guided operation is superior to the unguided operation from the efficiency and the success rate standpoints. Because guided trials were conducted after unguided trials, improvements may partially reflect increasing operator familiarity with the teleoperation interface rather than guidance alone. We hypothesize that the users were gaining more experience with the telerobotic platform from the unguided runs, which later benefited the guided operations. This does support the idea that increased experience with the telerobotic system can aid users for achieving a notable increase in success rate. Meanwhile, the divergence between the success rate and the mean duration indicates that the trained users are faster and more reliable. It further demonstrates that our telerobotic setup is effective once the teleoperators have undergone brief training.

To calculate the deviations in Figure~\ref{fig:5}B--E, we use the arc-length-aligned deviation method, which normalizes the timestamp of every trajectory to the relative completeness of the task, ranging from 0 to 1. The normalized trajectories are resampled at the pre-programmed frequency, followed by one-to-one comparisons between points at the same relative completeness, measured in both position and orientation, independent of time.

The first six unguided runs offer insights into how inexperienced users comprehend and perform path planning. The blue curves in Figure~\ref{fig:5} exhibit the average deviations of the unguided trajectories with respect to the multi-waypoint reference path in Figure~\ref{fig:5}B and the end-only path in Figure~\ref{fig:5}C. Comparing the blue curves in Figure~\ref{fig:5}B with the counterparts in Figure~\ref{fig:5}C, the smaller position deviations between the unguided path and the end-only reference suggest that inexperienced operators are willing to accept higher risks and tend to focus on the goals by simultaneously manipulating multiple DOFs. In Table~\ref{tab:1}, the average mean and max deviations condense the curves across the relative completeness into single numbers. The comparison of these average values shows that the unguided trajectories are more aligned with the end-only path in terms of position (\emph{x}, \emph{y}, \emph{z}). For orientation, as shown in Table~\ref{tab:1} and Figure S1 in the supplementary section, the unguided operations have equal similarity to the multi-waypoint and the end-only reference paths.

The multi-waypoint reference path (755.10 mm) is twice as long as the end-only path (375.04 mm). Yet, operators accomplished the task with almost identical path length (373.29 mm vs. 387.44 mm) and time (38.52 s vs. 42.78 s), when imitating the multi-waypoint reference. This suggests that the different reference paths might have inspired the users to find more efficient ways to achieve the goal. The pose at waypoint \#5, where the camera is perpendicular to the marker, is only one of the poses that achieve the goal. We found that users aligned the markers at different robot poses because varying camera orientations compensated for the different camera positions (\emph{x}, \emph{y}, \emph{z}). A similar trend was found in orientation deviations in Table~\ref{tab:1} and Figure S1; operators were able to limit the total orientation change when performing imitation. Additionally, the guided paths also exhibit reduced position and orientation deviations with respect to their corresponding references (see green vs. blue solid curves in Figure~\ref{fig:5}B and Figure~\ref{fig:5}C). This implies that the more efficient teleoperation paths were discovered under the influence of reference paths.

Interestingly, we also observe that the variance of path deviation is reduced in guided operations relative to unguided operations. As shown in Figure~\ref{fig:5}D and Figure~\ref{fig:5}E, the standard deviation of the path deviation of the unguided paths is larger at almost every single timestamp during the process. Meanwhile, with guidance, users were able to complete the task faster and with less time variance as suggested in Table~\ref{tab:1}.

These observations suggest that training with guidance is beneficial to the user in addition to increased practice. While these results may not suggest that a guided operation will always be faster, it suggests that users perform better when teleoperating with instructions. This is of critical importance in laboratory environments, which are cluttered or surrounded by costly equipment and operate under stringent safety constraints, demanding higher success rate and faster error rectification.

\begin{table*}[width=\textwidth,cols=8,pos=tp]
\caption{Statistics of the teleoperation performance. The path efficiency is the ratio of the reference path length to the teleoperation path length.}
\label{tab:1}
\centering
\footnotesize
\setlength{\tabcolsep}{3pt}
\renewcommand{\arraystretch}{1.2}
\begin{tabularx}{\textwidth}{@{}YYYYYYYY@{}}
\toprule
\multicolumn{8}{c}{\textbf{Position (mm)}} \\
\midrule
\textbf{Compared to} & \textbf{Human operation type} & \textbf{Average mean position deviation (\ensuremath{\downarrow})} & \textbf{Average max position deviation (\ensuremath{\downarrow})} & \textbf{Average path length (\ensuremath{\downarrow})} & \textbf{Reference path length} & \textbf{Average path efficiency (\ensuremath{\uparrow})} & \textbf{Duration (s) (\ensuremath{\downarrow})} \\
\midrule
Multi-waypoint path & Unguided & 79.11 & 163.24 & 598.46 & 755.10 & 1.52 & 77.39 \ensuremath{\pm} 43.22 \\
Multi-waypoint path & Guided (Multi-waypoint) & 65.53 & 157.85 & 373.29 & 755.10 & 2.23 & 38.52 \ensuremath{\pm} 15.57 \\
End-only path & Unguided & 72.07 & 131.62 & 598.46 & 375.04 & 0.76 & 77.39 \ensuremath{\pm} 43.22 \\
End-only path & Guided (End-only) & 56.78 & 113.32 & 387.44 & 375.04 & 1.02 & 42.78 \ensuremath{\pm} 17.96 \\
\bottomrule
\end{tabularx}
\par\medskip
\begin{tabularx}{\textwidth}{@{}YYYYYYY@{}}
\toprule
\multicolumn{7}{c}{\textbf{Orientation (rad)}} \\
\midrule
\textbf{Compared to} & \textbf{Human operation type} & \textbf{Average mean orientation deviation (\ensuremath{\downarrow})} & \textbf{Average max orientation deviation (\ensuremath{\downarrow})} & \textbf{Total orientation change (\ensuremath{\downarrow})} & \textbf{Reference path orientation change} & \textbf{Average path efficiency (\ensuremath{\uparrow})} \\
\midrule
Multi-waypoint path & Unguided & 1.43 & 3.08 & 14.74 & 17.37 & 1.35 \\
Multi-waypoint path & Guided (Multi-waypoint) & 0.92 & 2.88 & 11.03 & 17.37 & 1.72 \\
End-only path & Unguided & 1.42 & 3.14 & 14.74 & 16.12 & 1.25 \\
End-only path & Guided (End-only) & 1.35 & 3.12 & 10.75 & 16.12 & 1.57 \\
\bottomrule
\end{tabularx}
\par\smallskip\raggedright * Arrow (\ensuremath{\uparrow})(\ensuremath{\downarrow}) indicates the direction of change that corresponds to better performance.
\end{table*}

\begin{figure*}[pos=tp]
\centering
\includegraphics[width=1.0\textwidth,height=0.70\textheight,keepaspectratio]{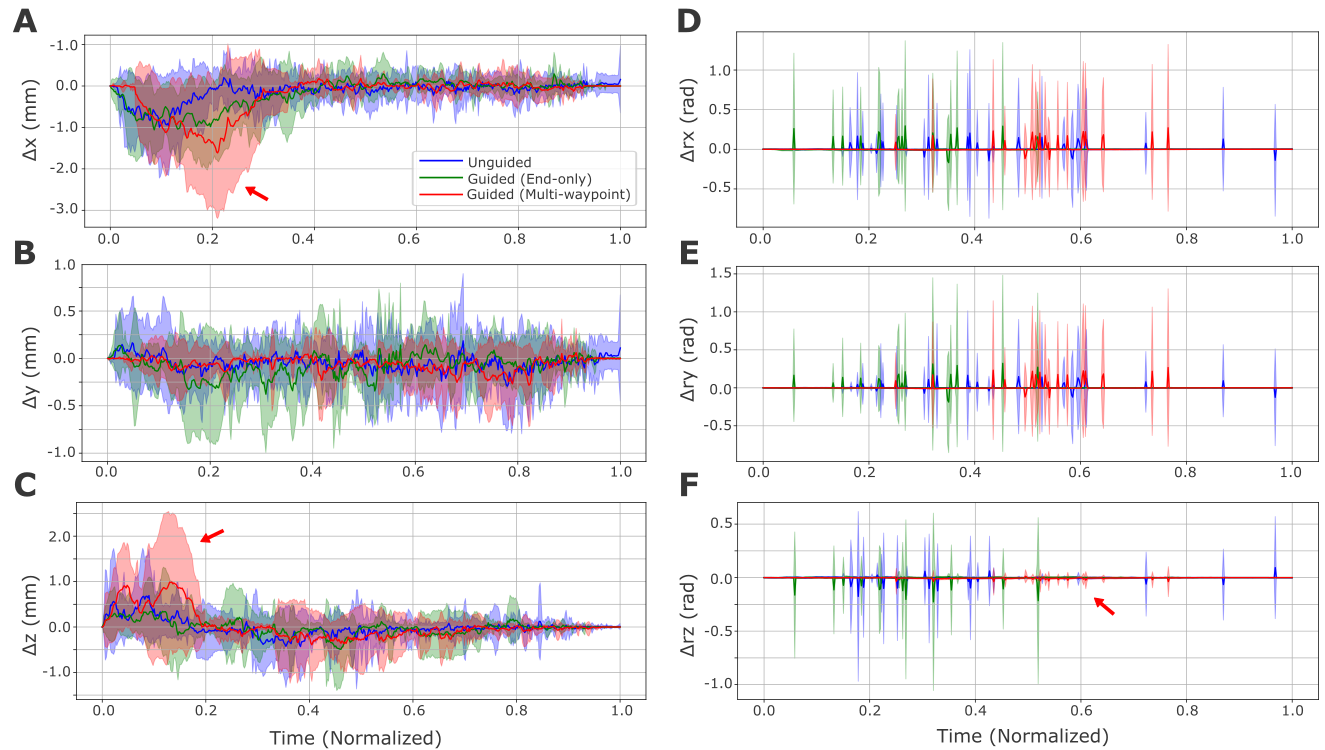}
\caption{Visualizing the average operation sequences in (A) \ensuremath{\Delta}x, (B) \ensuremath{\Delta}y, (C) \ensuremath{\Delta}z, (D) \ensuremath{\Delta}rx, (E) \ensuremath{\Delta}ry, and (F) \ensuremath{\Delta}rz for the unguided and the guided paths to understand the effects of guidance.}
\label{fig:6}
\end{figure*}

Although the multi-waypoint guided paths still deviate from the corresponding reference with some margins, as shown in Figure~\ref{fig:5}, the initial \(\mathrm{\Delta}x\) and \(\mathrm{\Delta}z\) spikes highlighted by the red arrows in Figure~\ref{fig:6}A and \ref{fig:6}C indicate that the operators attempted to imitate the multi-waypoint reference path from waypoint \#1 to \#3 shown in Figure~\ref{fig:4}. Similar evidence can be found from the orientation change; users mostly relied on pitch (\(\mathrm{\Delta}rx\)) and yaw (\(\mathrm{\Delta}ry\)) to align the camera during the multi-waypoint guided operation as nearly zero roll (\(\mathrm{\Delta}rz\)) is observed in Figure~\ref{fig:6}F indicated by the red arrow. In contrast, for the unguided and the end-only guided paths, operators primarily utilized all three motions of pitch, yaw, and roll during the first half of the task. In terms of position change, the unguided and the end-only guided paths are less distinguishable.

The operation sequences suggest that the inexperienced users tend to focus more on the objective and prefer to utilize frequent micro adjustments of all six DOFs over the entirety of the operation. The multi-waypoint reference shifts the participants' attention onto the operation sequence and educates them how to dissect a complex navigation into several segments with fewer simultaneous motions. As a result, users gain a better understanding of the correspondence between the controller and the robotic motions, leading to improved performance as shown in Table~\ref{tab:1}.

With the goal of making such reference paths simpler and easier to follow for the human operators, it is important to understand humans' tendency to search for the shortest route, but also recognize their limitations, \emph{e.g.}, lacking the ability to 1) deconvolve a complex path and 2) manipulate multiple DOFs simultaneously. A higher tolerance for imperfect operations should be considered to reduce collision risk, especially when designing for mission-critical tasks. Lastly, an intuitive feedback system, \emph{e.g.}, virtual/mixed reality~\cite{ref24,ref25}, can help operators visualize the reference path and reduce deviations.

To evaluate the performance of teleoperation within laboratory environments, our study focused on characterizing the general behaviors and limitations of operators under different guidance conditions. Because user performance inherently depends on factors such as trial experience and prior exposure to reference paths, future work should explore advanced evaluation frameworks tailored to the range of SDL applications~\cite{ref43}. As the SDL community continues to grow, it stands to benefit from established methodologies in more mature teleoperation domains, such as telesurgery~\cite{ref44,ref45,ref46}, where decades of research have yield robust protocols for benchmarking performance and safety.

\subsection{Understanding Collision and Singularity Risk}\label{understanding-collision-and-singularity-risk}

\begin{figure*}[pos=tp]
\centering
\includegraphics[width=0.86\textwidth,height=0.70\textheight,keepaspectratio]{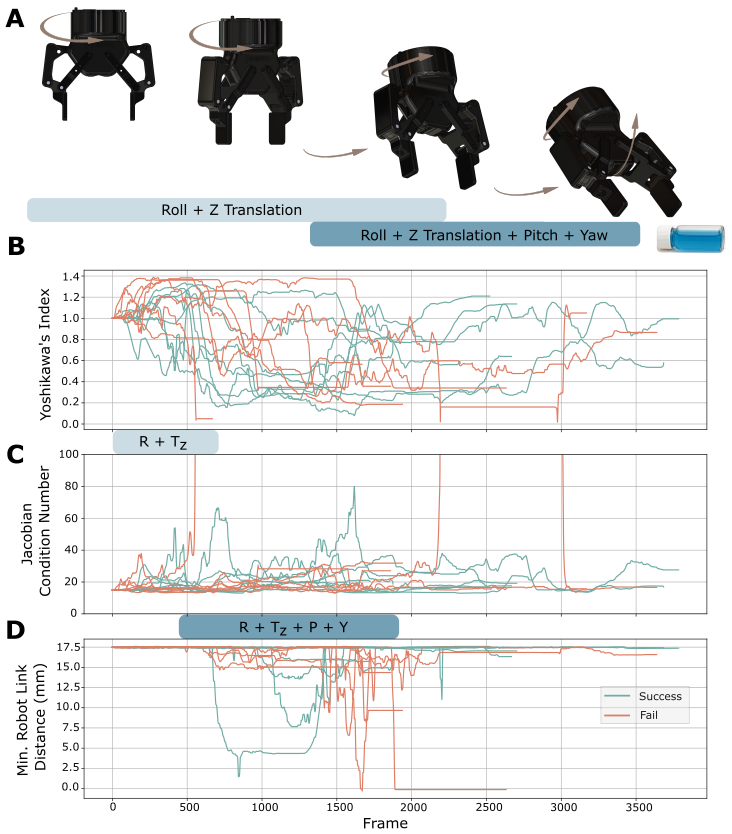}
\caption{Complex motions lead to a reduction in controllability and an increase in collision risk. (A) A series of complex gripper motions in the escalated fault scenario to pick up a dropped vial. The state of the UR3e arm is quantified in (B) Yoshikawa's index, (C) Jacobian condition number, and (D) minimum robot link distance, which quantify dexterity, proximity to singularities, and self-collision risk, respectively.}
\label{fig:7}
\end{figure*}

The failure modes most commonly observed during teleoperated SDL fault recovery include kinematic singularities, self-collisions, exceeding force limits, and arm overextensions. Mitigating such error states is a well-recognized challenge in teleoperation research, and an active area of research~\cite{ref32,ref33}. In real-world deployment, SDL teleoperation would likely be performed by operators without extensive robotics experience, making it essential to develop systems that not only simplify control but also aid users in avoiding high-risk configurations that may result in singularities and collisions~\cite{ref44,ref47}. In prior work, this has been achieved through the use of haptic feedback and visual cues~\cite{ref28,ref30}.

Although the primary focus of this study is the development and evaluation of teleoperation strategies for SDL fault recovery, our experiments in the earlier sections reveal that collisions and singularities can play a significant role in teleoperator performance. Preventing these anomalies is critical for SDLs, where robots interact with expensive, delicate, or hazardous materials and instruments. To better characterize these risks, we retrospectively analyzed all recorded teleoperation trials and introduced three quantitative metrics for assessment in Figure~\ref{fig:7}B--D. Yoshikawa's index \(\omega(q)\)~\cite{ref48} was used to quantify the dexterity of the current state of the robot:

\begin{equation}\label{eq:manipulability}
\omega(q) = \ \sqrt{\det(J(q)J^{T}(q))}
\end{equation}

where \(J(q)\) is the geometric Jacobian at the joint position \(q\); \(\omega(q) \rightarrow 0\) means approaching a singularity. The Jacobian condition number \(\kappa(J)\)~\cite{ref49} was evaluated to determine the robot's proximity to kinematic singularities, where small joint motions can lead to disproportionately large or ill-defined end-effector movements. It is the ratio of the largest \(\sigma_{\max}\) and the smallest \(\sigma_{\min}\) singular values of \(J\):

\begin{equation}\label{eq:condition-number}
\kappa(J) = \ \frac{\sigma_{\max}}{\sigma_{\min}}
\end{equation}

A large condition number (\emph{e.g.,} \textgreater100) indicates poor manipulability, approaching singularity. Finally, the minimum robot link distance \(d_{\min}\)~\cite{ref47} measures the risk of self-collision.

\begin{equation}\label{eq:link-distance}
d_{\min} = \ \min_{i,j}{dist(link_{i},link_{j})}
\end{equation}

When \(d_{\min}\) is small, the robot\textquotesingle s configuration is close to self-collision. Adjacent link pairs are excluded from the link distance calculation. Together, these metrics enable a systematic evaluation of collision and singularity risk during teleoperation.

As an illustrative example, the EF (escalated fault) scenario from Figure~\ref{fig:2}C involves a series of complex manipulations, most notably the grasping maneuver depicted in Figure~\ref{fig:7}A. This action requires precise motions to avoid wrist collision with nearby objects while achieving the correct alignment for vial retrieval. The retrospective analysis reveals that the robot experienced a period of low manipulability and high collision risk during the execution of this sequence between frames 500 and 2000 (Figure~\ref{fig:7}B). A few sudden spikes of the indices indicate that the self-collision and loss of manipulability caused the failures in these runs. These metrics provide a means for quantifying singularity and collision risks, which may be imperative for future implementations in SDLs that consider real-time evaluation, feedback, and risk mitigation during telerobotic operations.

Additionally, our teleoperation system includes a safety-oriented reset function that autonomously returns the robot to a neutral, untangled configuration under operator supervision. Note that collision avoidance during reset was not autonomously guaranteed and relied on workspace clearance and operator monitoring. Such features underscore the need for SDL teleoperation capabilities that support fault recovery while minimizing the risk of inducing new errors. Incorporating real-time monitoring and operator-assist mechanisms based on these metrics may further enhance the safety and robustness of teleoperation within future autonomous laboratory platforms.

\section{Real-World Demonstration of Telerobotic Operation for Laboratory Tasks}\label{real-world-demonstration-of-telerobotic-operation-for-laboratory-tasks}

\begin{figure*}[pos=tp]
\centering
\includegraphics[width=1.0\textwidth,height=0.70\textheight,keepaspectratio]{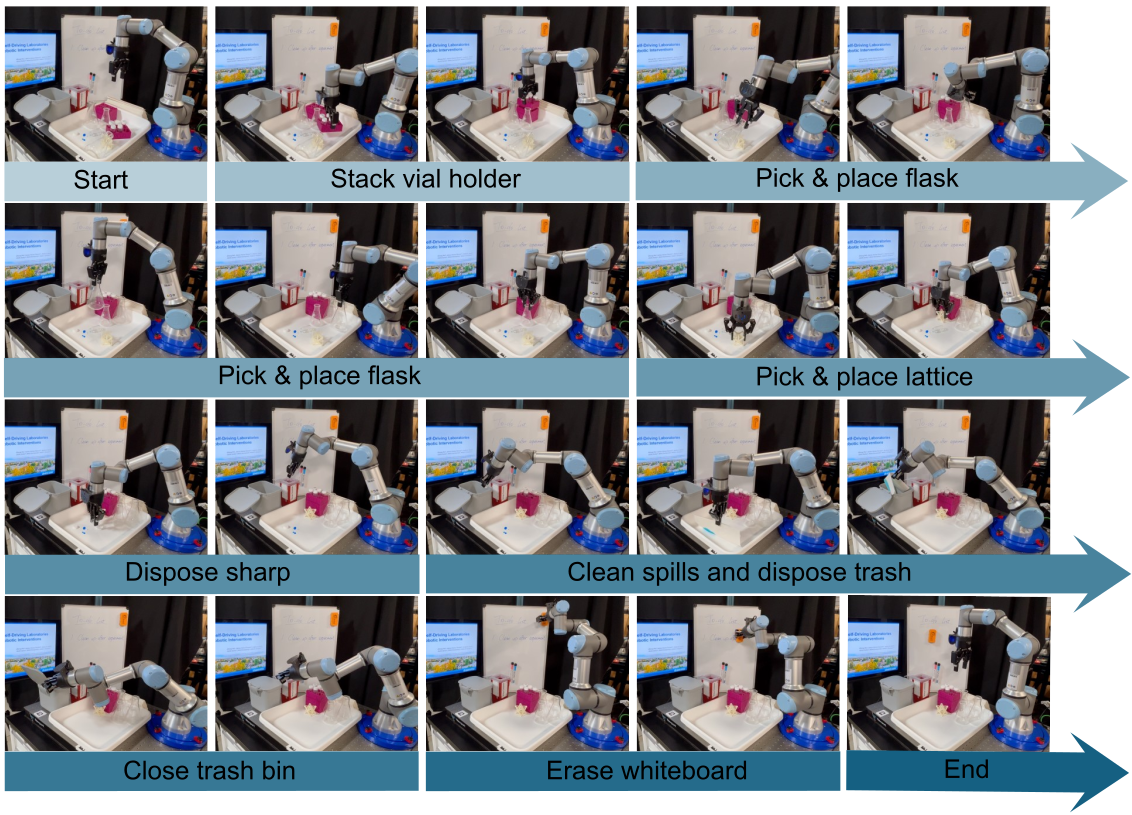}
\caption{A telerobotic janitor in action demonstrates that our setup can clean up after an autonomous experiment. The UR3e arm completed a series of lab tasks via teleoperation. See supplementary materials for the demo video S1.}
\label{fig:8}
\end{figure*}

While the earlier sections of this work focused on quantitatively evaluating teleoperation performance within controlled studies, we also sought to demonstrate the practical versatility of our system in realistic laboratory environments. This demonstration is intended as a qualitative illustration of task diversity rather than a controlled performance evaluation. To highlight the breadth of tasks achievable through teleoperation, we designed a set of real-world demonstrations using our telerobotic platform, referred to here as the ``Telerobotic Janitor'' (Figure~\ref{fig:8}). This demonstration illustrates that teleoperation provides a flexible and immediate means of performing diverse laboratory tasks, emphasizing its value for addressing fault conditions in SDLs that demand physical intervention.

Disruptive fault modes such as dropped samples, misplaced equipment, and incomplete task initialization are common in SDL operations and correspond closely to the PI, IE, and EF classes examined earlier in section~\ref{evaluating-control-strategies-for-telerobotic-operation}. To emulate the broad spectrum of such real-world challenges, we constructed a scenario involving a simulated chemical spill and charged the operator with restoring the laboratory environment to a condition suitable for resuming autonomous experimentation (Figure~\ref{fig:8}). Using only teleoperation, the operator successfully guided the UR3e robotic arm through a diverse series of tasks, including cleaning the simulated spill, organizing glassware and sample containers, disposing of regular and sharp waste, and erasing a written to-do list from a whiteboard. These tasks collectively represent the types of corrective and restorative actions required when SDLs encounter process interruptions, incorrect initializations, or escalated faults. Importantly, all tasks were completed without inducing additional errors or requiring direct human access to the workspace.

These demonstrations illustrate the general-purpose flexibility of our teleoperation framework and underscore its potential as a practical augmentation to SDLs. Beyond fault recovery, such telerobotic capabilities could support routine maintenance, reconfiguration of workflows, hazardous-material handling, and real-time human oversight, broadening the operational envelope of autonomous laboratory systems and strengthening their overall resilience.

\section{Conclusion}\label{conclusion}

In this work, we evaluated our teleoperation capability in an SDL platform to enable human intervention at various simulated system faults. Our results show that operators can rely solely on telerobotics to complete a series of laboratory tasks and restore automated workflows, and that their performance improves with experience --- highlighting the learnability and usability of the proposed system. We also assessed various controllability indices to characterize the risk of singularity and self-collision in teleoperation-mediated SDLs.

Through a comparative study of controller strategies, we found that effective interface design---particularly the decoupled control mode---can substantially improve success rates by 9.9\% and reduce task completion times by 25.6 s. Guided experiments additionally demonstrated that reference paths can help operators discover faster solutions, yet the observed discrepancies between executed trajectories and provided guidance indicate that reference path design must account for operator intent, strategy, and capability. These findings underscore the importance of human-centered design in teleoperation systems for SDL fault recovery.

Although the presented results demonstrate the feasibility of teleoperation-assisted fault recovery in SDL environments, several limitations should be acknowledged. First, the user studies were conducted with a relatively small number of primarily novice participants, which limits the generalizability of the observed performance trends. Second, the evaluated fault scenarios were intentionally simplified and represent controlled approximations of laboratory disruptions rather than the full complexity of real-world SDL operations. In the guidance study, guided trials were conducted after unguided trials, and therefore improvements in performance may partially reflect increasing operator familiarity with the teleoperation interface in addition to the effects of guidance itself. Furthermore, the ``Telerobotic Janitor'' demonstration was qualitative in nature and intended to illustrate operational flexibility rather than provide statistically rigorous benchmarking. The collision and singularity analyses were also performed retrospectively and were not integrated into a real-time assistive or autonomous control framework. Finally, while this work investigates teleoperation as a practical human-in-the-loop recovery mechanism, it does not demonstrate autonomous fault diagnosis, autonomous recovery planning, or shared-autonomy intervention. Addressing these limitations through larger-scale studies, more realistic laboratory workflows, counterbalanced experimental designs, and integrated assistive-control strategies represents an important direction for future research.

Overall, this work demonstrates that teleoperation can serve as a powerful mechanism for improving SDL resilience, offering practical guidelines for controller selection and reference path design. By enabling rapid and adaptive fault resolution within complex experimental workflows, teleoperation addresses key limitations of rigid automation. Collectively, these findings support the continued exploration of human-in-the-loop and shared-autonomy approaches for improving the operational robustness and resilience of future self-driving laboratory systems.

\FloatBarrier

\section*{Author Contributions}\label{author-contributions}

Z.W. and A.G. conceptualized the research and wrote the main manuscript text. A.S. and N.D. substantively revised the manuscript text. Z.W. performed data curation, investigation, formal analysis, and validation. Z.W., A.V.T., and A.F. prepared the figures and visualizations. B.G., E.D., C.S., and A.G. contributed to funding acquisition. A.G. led supervision and project administration. All authors reviewed and edited the manuscript.

\section*{Disclosure of AI Technology Usage}\label{disclosure-of-ai-technology-usage}

The authors acknowledge the use of OpenAI gpt-5.4 and gpt-5.5 models via LivChat, an LLNL internal AI tool, to assist with language editing, rephrasing and structuring for improving the clarity and flow of the manuscript. All AI-generated contents were reviewed, revised, and approved by the authors, who take full responsibility for the accuracy and integrity of the work.

\section*{Competing Interests}\label{competing-interests}

The authors declare no competing interests.

\section*{Funding}\label{funding}

This work was funded under the LLNL Laboratory Directed Research and Development project 25-ERD-032.

\section*{Acknowledgments}\label{acknowledgments}

This work was performed under the auspices of the U.S. Department of Energy by Lawrence Livermore National Laboratory under Contract DE-AC52--07NA27344. LLNL-JRNL-2012910

\bibliographystyle{model1-num-names}
\bibliography{references}
\end{document}


\let\WriteBookmarks\relax
\def\floatpagepagefraction{1}
\def\textpagefraction{.001}
\shorttitle{Supplementary material: Resilient SDLs through telerobotic interventions}
\shortauthors{Z. Wu et~al.}
\title[mode=title]{Supplementary Material\\[4pt] Toward Resilient Self-Driving Laboratories through Telerobotic Interventions}
\author[1]{Ziheng Wu}
\cormark[1]
\ead{wu57@llnl.gov}
\author[2,4]{Adrian Vecina Tercero}
\author[1]{Andre A. Fatehi}
\author[2]{Abhik Sarkar}
\author[3]{Brian Giera}
\author[3]{Eric B. Duoss}
\author[1]{Christopher M. Spadaccini}
\author[4]{Nikhil Deshpande}
\author[1]{Aldair E. Gongora}
\cormark[1]
\ead{gongora1@llnl.gov}
\affiliation[1]{organization={Materials Engineering Division, Lawrence Livermore National Laboratory},
  city={Livermore}, state={CA}, postcode={94550}}
\affiliation[2]{organization={Center for Applied Scientific Computing, Lawrence Livermore National Laboratory},
  city={Livermore}, state={CA}, postcode={94550}}
\affiliation[3]{organization={Engineering Directorate, Lawrence Livermore National Laboratory},
  city={Livermore}, state={CA}, postcode={94550}}
\affiliation[4]{organization={School of Computer Science, University of Nottingham},
  city={Nottingham}, country={UK}}
\cortext[1]{Corresponding authors.}

\renewcommand{\thefigure}{S\arabic{figure}}
\renewcommand{\thetable}{S\arabic{table}}

\makeatletter
\setlength{\@fptop}{0pt}
\setlength{\@fpsep}{10pt}
\setlength{\@fpbot}{0pt plus 1fil}
\setlength{\floatsep}{10pt}
\ExplSyntaxOn
\RenewDocumentCommand \maketitle { }
{
  \thispagestyle{first}
  \group_begin: \@title \group_end:
  \group_begin:
    \normalsize \stmauthors \par
    \stmcollab \par
    \footnotesize \itshape \stmaddress \par
  \group_end:
  \printFirstPageNotes
  \gdef\@pdfauthor{\infoauthors}
  \par\medskip
}
\ExplSyntaxOff
\makeatother
\maketitle
\begin{figure}[pos=!ht]
\centering
\includegraphics[width=1.0\textwidth,height=0.70\textheight,keepaspectratio]{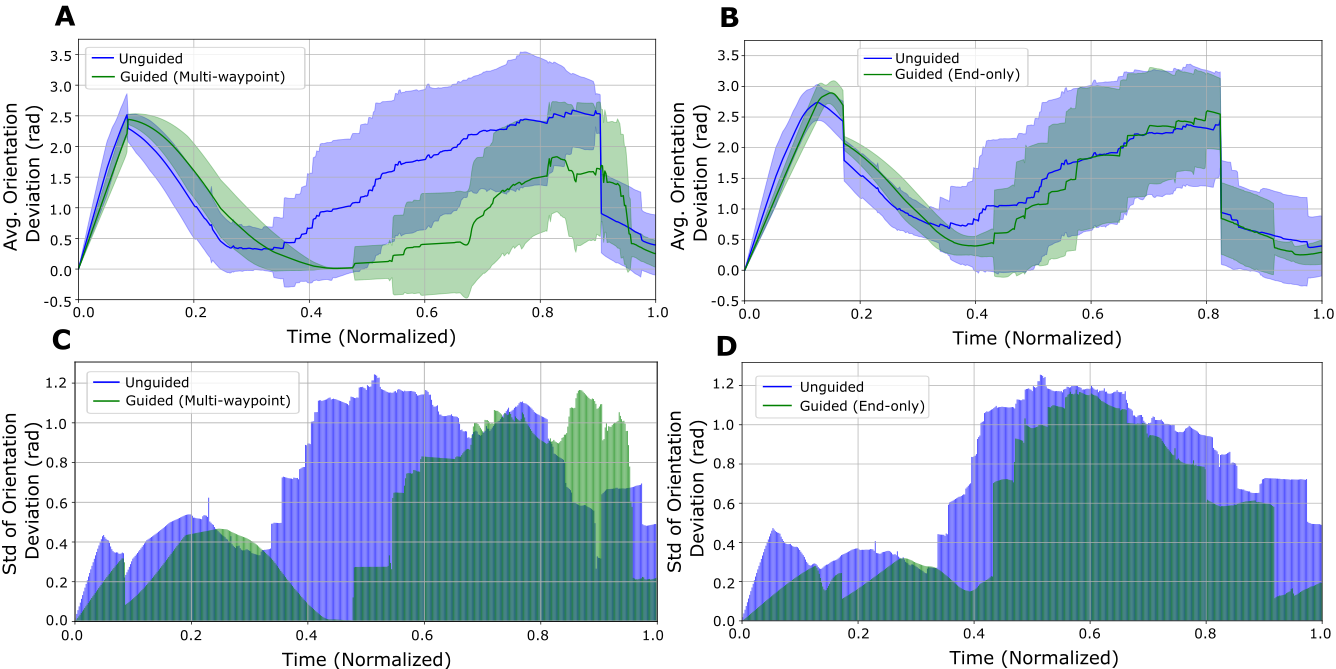}
\caption{Evaluating how practice and guidance affect users' performance and measuring the differences and similarities between the teleoperation paths and the reference paths. Quantifying (A) the orientation deviations and (C) the corresponding variations by comparing the unguided and multi-waypoint guided paths with the multi-waypoint reference, as well as (B) the orientation deviations and (D) the corresponding variations by comparing the unguided and end-only guided paths with the end-only reference. The solid curves in (A) and (B) track the average deviations, and the shaded areas present \ensuremath{\pm} 1 standard deviation which is represented in (C) and (D). Note that only successful trials contribute to the deviation calculation.}
\label{fig:S1}
\end{figure}

\begin{table}[width=\textwidth,cols=8,pos=!ht]
\caption{Statistical summary of controller comparison results between coupled and decoupled teleoperation modes. Continuous metrics were computed from participant-level paired averages using successful trials only. Bootstrap confidence intervals were computed using 10,000 resamples. Success-rate confidence intervals were computed using Wilson intervals. Because of the exploratory nature and limited participant count of the study, statistical analyses are intended to characterize performance trends and uncertainty rather than establish definitive generalizable conclusions.}
\label{tab:S1}
\centering
\footnotesize
\setlength{\tabcolsep}{3pt}
\renewcommand{\arraystretch}{1.1}
\begin{tabularx}{\textwidth}{@{}ZYYYYYYY@{}}
\toprule
\textbf{Metric} & \textbf{Coupled mean \ensuremath{\pm} SD} & \textbf{Decoupled mean \ensuremath{\pm} SD} & \textbf{Mean paired difference (Decoupled \ensuremath{-} Coupled)} & \textbf{Bootstrap 95\% CI} & \textbf{Wilcoxon p-value} & \textbf{Paired t-test p-value} & \textbf{Cohen's dz} \\
\midrule
Success rate & 70.50\% & 80.40\% & \textemdash & \textemdash & \textemdash & \textemdash & \textemdash \\
Duration (s) & \textemdash & \textemdash & $-25.624$ & $-36.584$ to $-14.223$ & $0.031$ & $0.011$ & $-1.597$ \\
Path length (mm) & \textemdash & \textemdash & $-123$ & $-284$ to $56$ & $0.312$ & $0.259$ & $-0.52$ \\
Rotation (rad) & \textemdash & \textemdash & $-3.99$ & $-6.585$ to $-1.224$ & $0.094$ & $0.046$ & $-1.078$ \\
\bottomrule
\end{tabularx}
\end{table}

\begin{table}[width=\textwidth,cols=6,pos=!ht]
\caption{Exploratory statistical summary of guidance-study comparisons between the multi-waypoint guided condition and unguided operation. Continuous metrics were computed from participant-level paired averages using successful trials only. Bootstrap confidence intervals were computed using 10,000 resamples.}
\label{tab:S2}
\centering
\footnotesize
\setlength{\tabcolsep}{3pt}
\renewcommand{\arraystretch}{1.1}
\begin{tabularx}{\textwidth}{@{}>{\hsize=1.45\hsize}Z>{\hsize=.95\hsize}Y>{\hsize=.95\hsize}Y>{\hsize=.95\hsize}Y>{\hsize=1.0\hsize}Y>{\hsize=.70\hsize}Y@{}}
\toprule
\textbf{Metric} & \textbf{Unguided mean} & \textbf{Guided (multi-waypoint) mean} & \textbf{Mean paired difference (Guided \ensuremath{-} Unguided)} & \textbf{Bootstrap 95\% CI} & \textbf{Wilcoxon p-value} \\
\midrule
Success rate & 64.30\% {[}95\% CI: 49.2 \% to 77.0 \%{]} & 76.20\% {[}95\% CI: 54.9 \% to 89.4 \%{]} & \textemdash & \textemdash & \textemdash \\
Duration (s) & \textemdash &  & $-40.087$ & $-55.935$ to $-26.265$ & $0.016$ \\
Path length (mm) & \textemdash & \textemdash & $-218$ & $-365$ to $-81$ & $0.047$ \\
Mean position deviation (mm) & \textemdash & \textemdash & $-13$ & $-27$ to $2$ & $0.156$ \\
Maximum position deviation (mm) & \textemdash & \textemdash & $-38$ & $-62$ to $-15$ & $0.047$ \\
Mean orientation deviation (rad) & \textemdash & \textemdash & $-0.563$ & $-0.723$ to $-0.409$ & $0.016$ \\
Maximum orientation deviation (rad) & \textemdash & \textemdash & $-0.187$ & $-0.309$ to $-0.078$ & $0.016$ \\
Total orientation change (rad) & \textemdash & \textemdash & $-3.384$ & $-7.068$ to $0.067$ & $0.156$ \\
Path efficiency & \textemdash & \textemdash & $0.738$ & $0.175$ to $1.308$ & $0.078$ \\
Orientation efficiency & \textemdash & \textemdash & $0.374$ & $0.146$ to $0.616$ & $0.047$ \\
\bottomrule
\end{tabularx}
\end{table}

\begin{table}[width=\textwidth,cols=6,pos=!ht]
\caption{Exploratory statistical summary of guidance-study comparisons between the end-waypoint guided condition and unguided operation. Continuous metrics were computed from participant-level paired averages using successful trials only. Bootstrap confidence intervals were computed using 10,000 resamples.}
\label{tab:S3}
\centering
\footnotesize
\setlength{\tabcolsep}{3pt}
\renewcommand{\arraystretch}{1.1}
\begin{tabularx}{\textwidth}{@{}>{\hsize=1.45\hsize}Z>{\hsize=.95\hsize}Y>{\hsize=.95\hsize}Y>{\hsize=.95\hsize}Y>{\hsize=1.0\hsize}Y>{\hsize=.70\hsize}Y@{}}
\toprule
\textbf{Metric} & \textbf{Unguided mean} & \textbf{Guided (end-waypoint) mean} & \textbf{Mean paired difference (Guided \ensuremath{-} Unguided)} & \textbf{Bootstrap 95\% CI} & \textbf{Wilcoxon p-value} \\
\midrule
Success rate & 64.30\% {[}95\% CI: 49.2 \% to 77.0 \%{]} & 66.67\% {[}95\% CI: 45.4 \% to 82.8 \%{]} & \textemdash & \textemdash & \textemdash \\
Duration (s) & \textemdash & \textemdash & $-40.957$ & $-58.261$ to $-20.460$ & $0.031$ \\
Path length (mm) & \textemdash & \textemdash & $-227$ & $-347$ to $-99$ & $0.031$ \\
Mean position deviation (mm) & \textemdash & \textemdash & $-24$ & $-31$ to $-15$ & $0.016$ \\
Maximum position deviation (mm) & \textemdash & \textemdash & $-53$ & $-66$ to $-42$ & $0.016$ \\
Mean orientation deviation (rad) & \textemdash & \textemdash & $-0.188$ & $-0.470$ to $-0.107$ & $0.375$ \\
Maximum orientation deviation (rad) & \textemdash & \textemdash & $-0.003$ & $-0.106$ to $-0.088$ & $1$ \\
Total orientation change (rad) & \textemdash & \textemdash & $-3.100$ & $-9.920$ to $1.183$ & $0.156$ \\
Path efficiency & \textemdash & \textemdash & $-0.389$ & $-0.782$ to $-0.060$ & $0.078$ \\
Orientation efficiency & \textemdash & \textemdash & $0.198$ & $-0.089$ to $0.457$ & $0.219$ \\
\bottomrule
\end{tabularx}
\end{table}